\documentclass{article}

\usepackage[dblblindworkshop, final]{neurips_2025}

\usepackage[utf8]{inputenc} % allow utf-8 input
\usepackage[T1]{fontenc}    % use 8-bit T1 fonts
\usepackage{hyperref}       % hyperlinks
\usepackage{url}            % simple URL typesetting
\usepackage{booktabs}       % professional-quality tables
\usepackage{amsfonts}       % blackboard math symbols
\usepackage{nicefrac}       % compact symbols for 1/2, etc.
\usepackage{microtype}      % microtypography
\usepackage{xcolor}         % colors
\usepackage{graphicx}
\usepackage{multirow} 
\usepackage{amsmath}
\usepackage{svg}
\usepackage{breqn}

\workshoptitle{Efficient Reasoning}

\title{Thinking in Many Modes: How Composite Reasoning Elevates Large Language Model Performance with Limited Data}

\author{%
 Zishan Ahmad \\%\thanks{Use footnote for providing further information
  PhiLabs, Quantiphi Inc\\
  Bengaluru, India \\
  \texttt{zishan.ahmad@quantiphi.com} \\
  \And
  Saisubramaniam Gopalakrishnan \\
  PhiLabs, Quantiphi Inc\\
  Bengaluru, India \\
  \texttt{gopalakrishnan.saisubramaniam@quantiphi.com}
}

\begin{document}

\maketitle

\begin{abstract}
Large Language Models (LLMs), despite their remarkable capabilities, rely on singular, pre-dominant reasoning paradigms, hindering their performance on intricate problems that demand diverse cognitive strategies.
To address this, we introduce Composite Reasoning (CR), a novel reasoning approach empowering LLMs to dynamically explore and combine multiple reasoning styles like deductive, inductive, and abductive for more nuanced problem-solving.     Evaluated on scientific  and medical question-answering benchmarks, our approach outperforms existing baselines like Chain-of-Thought (CoT) and also surpasses the accuracy of DeepSeek-R1 style reasoning (SR) capabilities, while demonstrating superior sample efficiency and adequate token usage. Notably, CR adaptively emphasizes domain-appropriate reasoning styles. It prioritizes abductive and deductive reasoning for medical question answering, but shifts to causal, deductive, and inductive methods for scientific reasoning. 
Our findings highlight that by cultivating internal reasoning style diversity, LLMs  acquire more robust, adaptive, and efficient problem-solving abilities.
\end{abstract}

\section{Introduction}
\label{sec:intro}
The remarkable progress in Large Language Models (LLMs) has led to significant advancements in natural language understanding and generation, largely due to their enhanced reasoning capabilities. While traditional approaches often rely on a single dominant reasoning paradigm, we hypothesize that enabling LLMs to leverage a diverse repertoire of reasoning strategies—such as deductive, inductive, abductive, and decompositional reasoning—can lead to more robust and accurate performance, particularly on complex tasks.%\textcolor{red}{Suggestion - Since we are just beginning the intro section, it would be better to avoid words such as 'we hypothesize' etc. in the very beginning}

Recent state-of-the-art LLMs often employ Chain-of-Thought (CoT) prompting \cite{wei2022chain}, which elicits a step-by-step reasoning process. While effective, this process frequently exhibits a predominantly deductive flow. Our work, however, builds on the understanding that real-world problems demand greater flexibility, potentially requiring models to dynamically synthesize different reasoning methods. Previous efforts have explored enhancing LLM reasoning through improved decoding \cite{wang2022self}, incorporating external knowledge \cite{liu2023pre}, or using techniques like self-consistency \cite{naik2023diversity}. However, there has been limited exploration of explicitly encouraging models to internally explore and integrate multiple distinct reasoning strategies during problem-solving.

To address this gap, we propose a novel composite reasoning (CR) approach. Our method encourages LLMs to explore and combine different reasoning strategies, allowing the model to consider multiple perspectives and pathways for more accurate and well-supported answers. We evaluate this approach using parameter-efficient fine-tuning (PEFT) techniques like LoRA \cite{hu2022lora} and Group Relative Policy Optimization (GRPO) \cite{shao2024deepseekmath} to assess performance in resource-constrained settings.

We conduct extensive experiments on three challenging and diverse datasets: MedMCQA \cite{pal2022medmcqa}, MedXpertQA \cite{zuo2025medxpertqa}, and ARC-Complex \cite{clark2018think}. All fine-tuning and training were conducted using a maximum of 1,500 samples per dataset. We compare our CR strategy against standard CoT and a Standard Reasoning (SR) approach, demonstrating a compelling performance advantage. Furthermore, we show that GRPO with an outcome-based reward function (based solely on answer correctness) allows our CR approach to implicitly foster a more flexible and multi-faceted reasoning process that adapts to the specific demands of each domain.

We summarize our key contributions as follows:
\textit{(i).} A novel composite-reasoning approach that encourages LLMs to explore and adapt multiple reasoning strategies,  
\textit{(ii).} We demonstrate the effectiveness of this approach on three challenging datasets within a resource-constrained training setting (maximum 1,500 samples), highlighting its superior sample efficiency, 
\textit{(iii).} We show that GRPO with an outcome-based reward effectively guides our CR approach to explore and tailor diverse reasoning strategies to domain-specific needs, \textit{and}
\textit{(iv).} Our results indicate significant performance improvements over standard CoT and SR baselines, highlighting the benefits of CR in terms of accuracy and token effectiveness in resource-constrained scenarios.

\section{Methodology}
\label{sec:methodology}

This section details our experimental framework, which investigates the performance of our Composite Reasoning (CR) approach under resource constraints. All fine-tuning and training stages used a maximum of 1,500 samples from the official training splits of each dataset. We evaluate our models on the official test sets of ARC-Complex (1,119 questions), MedMCQA (4,183 questions) and MedXpertQA (950 questions).

We investigate three distinct reasoning paradigms for which initial trajectories were generated using a base Qwen-2.5-7B-Instruct model (except for SR, which was sourced from Deepseek-r1-7B \cite{guo2025deepseek}). %\textcolor{red}{Suggestion - should be fine to use simple present tense for the above two paras here - past makes the text sound very complex while reading}

% \begin{figure*}
%     \centering
%     \includesvg[width=0.8\linewidth]{medmcqa_reason_eg.svg}
%     \caption{\small Example of Composite Reasoning style for a MedMCQA Question}
%     \label{fig:egcr}
% \end{figure*}

\begin{figure*}
    \centering
    \includegraphics[width=\linewidth]{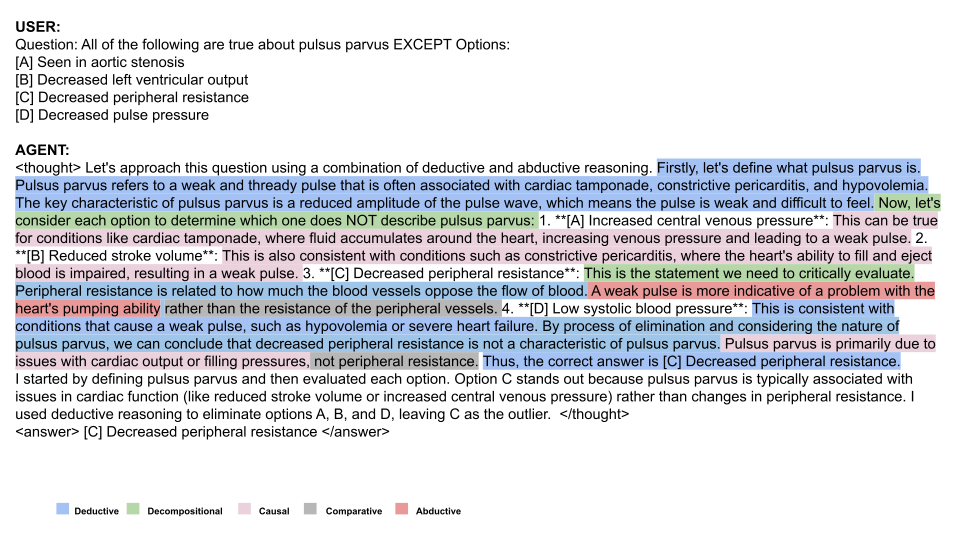}
    \caption{\small Example of Composite Reasoning style for a MedMCQA Question}
    \label{fig:egcr}
\end{figure*}
\subsection{Supervised Fine-Tuning (SFT) with LoRA}
\label{ssec:sft_lora}

\begin{enumerate}
    \item \textbf{Chain-of-Thought (CoT)} \cite{wei2022chain}:
    This is the conventional method for eliciting sequential reasoning. The intuition is to prompt the model to ``think step-by-step'', which often leads to a logical, deductive-like progression that can improve accuracy on complex tasks.

    \item \textbf{Standard Reasoning (SR):}
    This baseline uses high-quality, pre-generated reasoning trajectories from Deepseek-r1-7B model. % \textcolor{red}{ possible to name/reference the model?}.  
    The purpose here is to test whether distilling a highly-polished reasoning style from a powerful external source is an effective fine-tuning strategy, even in a low-data setting.

    \item \textbf{Composite Reasoning (CR):}
    Our approach explicitly prompts the model to dynamically explore and synthesize diverse reasoning strategies. The intuition behind this is to move beyond a single, linear thought process. It encourages the model to leverage a full ``toolkit'' of reasoning, including hypothesis generation (abduction), generalization (induction), and logical breakdowns (decomposition), thereby making it more adaptable to a wider range of problems.
    An illustrative example of a CR-generated thought process, showcasing these characteristics with annotated reasoning styles, is presented in Figure \ref{fig:egcr}.
\end{enumerate}
%\textcolor{red}{We can use phrases like "We finetune.." instead of "The base LLM underwent.." -- can again stick to simple present for this entire portion} 
We finetune the base LLM using Supervised Fine-Tuning (SFT) using Low-Rank Adaptation (LoRA) on these generated trajectories. The intuition of SFT is to teach the model to imitate the reasoning styles we curated for each paradigm. This process essentially instills the desired ``thinking patterns'' (CoT, SR, or CR) into the model's behavior. Following SFT, we applied Group Relative Policy Optimization (GRPO) \cite{shao2024deepseekmath}, a reinforcement learning algorithm tailored for scenarios with sparse rewards. The core intuition of using GRPO with an outcome-based reward is to let the model self-refine its reasoning process based on a simple, yet powerful, signal: whether the final answer is correct or not. This encourages the model to generate reasoning that is not just plausible, but pragmatically effective at solving the task, without needing complex human-in-the-loop reward modeling for each reasoning step.

\begin{table*}[t]
\centering
\scriptsize
\begin{tabular}{llcccccc}
\hline
\multirow{2}{*}{\textbf{Method}} 
  & \multirow{2}{*}{\textbf{Reasoning}} 
    & \multicolumn{2}{c}{\textbf{ARC-C}} 
      & \multicolumn{2}{c}{\textbf{MedMCQA}}
       & \multicolumn{2}{c}{\textbf{MedXpertQA}}\\
  &  
    & \textbf{Accuracy} 
      & \textbf{Avg Token Length} 
        & \textbf{Accuracy} 
          & \textbf{Avg Token Length} 
           & \textbf{Accuracy} 
          & \textbf{Avg Token Length}\\
\hline

\multirow{4}{*}{Prompt} 
  & Direct       & 60.06\%      & 38   & 45.88\%      & 42  & 5\% & 35\\
  & CoT          & 73.7\%       & 254  & 52.77\%      & 205 & 8\% & 287\\
  & SR$^\dagger$ & 80.54\%      & 515  & 48.96\%      & 619 & 7.9\% & 1,139\\
  & CR           & 83.10\%      & 316  & 54.62\%      & 335 & 7.8\% & 398\\
\hline

\multirow{3}{*}{SFT} 
  & CoT          & 92.31\%      & 252  & 55.35\%      & 207 & 14.34\% & 282 \\
  & SR           & 75.20\%      & 518  & 50.96\%      & 612 & 11\% & 1,146\\
  & CR           & 92.22\%      & 320  & 55.20\%      & 338 & 14.1\% & 405\\
\hline

\multirow{3}{*}{GRPO} 
  & CoT          & 89.5\%       & 249  & 55.10\%      & 189 & 13.1\% & 412\\
  & SR$^\dagger$ & 78.42\%      & 524  & 49.23\%      & 601 & 9.1\% & 1,110\\
  & CR & 90.35\%      & 331  & 55.84\%      & 331 & 10.08\% & 426 \\ 
\hline

\multirow{3}{*}{SFT + GRPO} 
  & CoT          & 93.85\%      & 247  & 55.74\%      & 208 & 14.63\% & 479\\
  & SR           & 80.20\%      & 518  & 51.80\%      & 617 & 11.47\% & 1,112\\
  & CR           & \textbf{94.99\%} & 339 & \textbf{56.30\%} & 313 & 15.9\% & 549\\
\hline
\end{tabular}
\caption{\small\label{tab:main_results}
Exact-Match Accuracy (\%) on ARC-Complex, MedMCQA and MedXpertQA datasets across reasoning strategies and methods, with output token lengths. Best in \textbf{bold}. $^\dagger$ experiments on \textit{Deepseek-r1-7B}; others on \textit{Qwen2.5-7B-Instruct}.
}
\end{table*}

\begin{table}[t]
\centering
\small

\begin{tabular}{cc}
\hline
\textbf{Model}                     & \textbf{MedMCQA Acc. (\%)} \\
\hline
BioMistral-7B                       & 40.2   \\
OpenBioLLM-8B                       & 54.1   \\
ChatDoctor                          & 31.5   \\
PMC-LLaMA-7B                        & 29.8   \\
Baize-Healthcare                    & 31.3   \\
MedAlpaca-7B                        & 32.9   \\
Meditron-7B                         & 31.1   \\
PMC-LLaMA-13B                       & 37.7   \\
MedAlpaca-13B                       & 35.7   \\
ClinicalCamel                       & 45.8   \\
Huatuo-7B                           & 24.8 \\
HuatuoGPT-o1-8B                     & \textbf{60.4}   \\
HuatuoGPT-o1-8B w/o RL                     & 57.9   \\
UltraMedical-8B                     & 58.3   \\
\hline
CR Prompt                           & 54.62  \\
CR SFT                              & 55.20  \\
CR GRPO                             & 55.84  \\
CR SFT + GRPO                       & \textbf{56.30} \\
\hline
\end{tabular}
\caption{ \small MedMCQA accuracy (\%) comparison with existing baseline models from the 7-13B parameter size category. Note that CR is trained on only 1,500 training samples as compared to HuatuoGPT-o1 (40k samples) and UltraMedical (410k samples), yet remains competitive.}
\label{tab:medmcqa_all_models}
\end{table}
% \vspace{-2em}
% \subsubsection{Evaluation Metrics}
% \label{ssec:evaluation_metrics}

% The primary evaluation metric across all tasks and models was \textbf{exact match accuracy}. A predicted answer was considered correct if and only if the text within the \texttt{<answer> </answer>} tags precisely matched the ground truth answer string.

\section{Results and Analysis}
\label{sec:results_analysis}

%This section presents our empirical results on the ARC-Complex (ARC-C) and MedMCQA datasets, focusing on the performance of our Composite Reasoning (CR) approach against Chain-of-Thought (CoT) and Standard Reasoning (SR) baselines.
We present our empirical results on the ARC-Complex (ARC-C), MedMCQA, and MedXpertQA datasets. We analyze the performance of our proposed Composite Reasoning (CR) approach against Chain-of-Thought (CoT) and Standard Reasoning (SR) baselines. The detailed accuracy scores and average token lengths are presented in Table \ref{tab:main_results}. In the direct zero-shot prompting setting, our Composite Reasoning (CR) prompt consistently outperforms standard Direct Prompting and CoT across all three datasets. While CR is slightly outperformed by SR on the highly complex MedXpertQA dataset (7.8\% vs. 7.9\%), this initial advantage on the other two datasets demonstrates the effectiveness of our prompt in eliciting stronger baseline reasoning. As indicated by the low overall accuracy scores, MedXpertQA is a significantly more difficult task requiring a higher level of domain-specific reasoning, which is reflected in the extremely verbose nature of the SR model's initial trajectories on this dataset (1,139 average tokens).

Supervised Fine-Tuning (SFT) using only 1,500 samples substantially enhances all strategies. Notably, CR SFT and CoT SFT achieve strong, competitive results, significantly outperforming SR SFT on all three datasets. %This suggests that in our low-data, parameter-efficient setting, fine-tuning on self-generated CR or CoT paths is still effective than attempting to distill reasoning from an external model.
The addition of GRPO to the SFT-tuned models consistently yields the highest performance. The CR SFT + GRPO configuration achieves the highest accuracy on both ARC-C (94.99\%) and MedMCQA (56.30\%), and it secures the top performance on MedXpertQA (15.9\%). This demonstrates the potent synergy of CR with SFT and subsequent outcome-based reward tuning, enabling the model to explore and optimize reasoning paths better than CoT and SR to reach peak performance with limited training data. This synergy is particularly evident when analyzing the performance gain on the challenging MedXpertQA dataset. On this task, our CR method achieves a substantial gain of 8.1\% (from 7.8\% to 15.9\%), which is significantly larger than the gains of CoT (6.63\%) and SR (3.57\%). This contrasts with the MedMCQA dataset, where the gain is more modest (1.68\%), suggesting that CR's ability to learn from limited data is most pronounced when the problem requires deep, non-memorization-based reasoning.

Analysis of reasoning chain lengths reveals a compelling accuracy-verbosity trade-off. As noted, SR produces the longest reasoning paths, but this verbosity does not translate to higher accuracy in our fine-tuning setup. While CoT is generally the most concise, CR strikes a better balance, achieving superior accuracy with moderately longer but more effective reasoning chains. For the highly complex MedXpertQA dataset, the average token counts for both CR and CoT increase after GRPO training (from 405 to 549 for CR, and 282 to 479 for CoT), indicating that the models are generating more detailed reasoning to solve the harder problems. This suggests that GRPO optimizes CR towards more token-effective reasoning on simpler tasks while encouraging necessary verbosity for complex ones.

The sample efficiency of our method is particularly noteworthy on the MedMCQA dataset. As shown in Table \ref{tab:medmcqa_all_models}, our CR SFT + GRPO model achieves an accuracy of 56.30\%. This is highly competitive with medical LLMs like HuatuoGPT-o1-8B and UltraMedical-8B, despite our model being trained on only 1,500 samples, a small fraction of the 40k and 410k domain specific samples, respectively, used by those baselines. This highlights the remarkable sample efficiency of our CR approach.
The analysis of reasoning style dynamics (visualized in Figures \ref{fig:arcanalysis} and \ref{fig:medanalysis} in the Appendix) reveals that GRPO selectively modifies the problem-solving approaches of our models in a domain-dependent manner. For a more detailed discussion of these stylistic shifts, refer to Appendix \ref{subsec:appanalysis}.

\section{Conclusion}
\label{sec:conclude}
In this work, we introduce Composite Reasoning (CR), a method that enhances LLMs’ complex reasoning by encouraging the exploration and integration of diverse strategies. In a resource-constrained (1,500-sample) LoRA-based fine-tuning setup on challenging datasets, including ARC-Complex, MedMCQA, and the highly demanding MedXpertQA, our CR approach consistently outperforms standard Chain-of-Thought (CoT) and Standard Reasoning (SR) baselines. This performance is particularly noteworthy on the most difficult tasks, where CR, combined with GRPO-based fine-tuning, achieves a significantly greater performance gain than other methods. Our experiments underscore CR's remarkable sample efficiency, allowing it to compete with domain-specific LLMs on MedMCQA despite using orders of magnitude less training data. By encouraging diverse reasoning strategies like deductive, inductive, abductive, etc., our findings show that LLMs can develop more robust, adaptive, and effective problem-solving skills.

%As a future work, it would be interesting to study the effects of process rewards to encourage different reasoning styles in different domains.

% \bibliographystyle{IEEEtran}  
% \bibliographystyle{plainnat}
% %plainnat,abbrvnat,unsrtnat
% \small
% \bibliography{Reference}
% % \bibliographystyle{IEEEtran}
% \normalsize
% \section*{References}

% References follow the acknowledgments in the camera-ready paper. Use unnumbered first-level heading for
% the references. Any choice of citation style is acceptable as long as you are
% consistent. It is permissible to reduce the font size to \verb+small+ (9 point)
% when listing the references.
% Note that the Reference section does not count towards the page limit.
% \medskip

% {
% \small

% [1] Alexander, J.A.\ \& Mozer, M.C.\ (1995) Template-based algorithms for
% connectionist rule extraction. In G.\ Tesauro, D.S.\ Touretzky and T.K.\ Leen
% (eds.), {\it Advances in Neural Information Processing Systems 7},
% pp.\ 609--616. Cambridge, MA: MIT Press.

% [2] Bower, J.M.\ \& Beeman, D.\ (1995) {\it The Book of GENESIS: Exploring
%   Realistic Neural Models with the GEneral NEural SImulation System.}  New York:
% TELOS/Springer--Verlag.

% [3] Hasselmo, M.E., Schnell, E.\ \& Barkai, E.\ (1995) Dynamics of learning and
% recall at excitatory recurrent synapses and cholinergic modulation in rat
% hippocampal region CA3. {\it Journal of Neuroscience} {\bf 15}(7):5249-5262.
% }

%%%%%%%%%%%%%%%%%%%%%%%%%%%%%%%%%%%%%%%%%%%%%%%%%%%%%%%%%%%%

\appendix

\section{Appendix}

\subsection{Experimental Setup}
\label{ssec:experimental_setup}

All experiments were conducted on a single NVIDIA A100 80GB GPU. All our experiments are based on Qwen-2.5-7B and Deepseek-r1-7B models both containing around 7 billion parameters. Each SFT training took around 7 hours on a single GPU, while GRPO tuning varied took between 24-48 hours for different experiments. We employed a consistent configuration across both the Supervised Fine-Tuning (SFT) and Generalized Reinforcement Preference Optimization (GRPO) phases. LoRA adapters were configured with a rank \( r = 32 \) and an alpha \( \alpha = 64 \), resulting in a scaling factor \( s = \alpha / r = 2 \). The target modules for LoRA integration included the following linear layers: \texttt{q\_proj}, \texttt{k\_proj}, \texttt{v\_proj}, \texttt{o\_proj}, \texttt{gate\_proj}, \texttt{up\_proj}, and \texttt{down\_proj}. These modules correspond to the query, key, value, and output projection layers in the attention blocks, as well as the feed-forward network components within the transformer architecture.
During the SFT phase, we used a learning rate of $10^{-4}$, a batch size of 8, and trained for 12 epochs. The optimizer employed was AdamW with a weight decay of 0.001, and the learning rate scheduler followed a linear warmup with decay strategy.
In the GRPO phase, a new LoRA adapter was trained using a learning rate of $10^{-4}$, a batch size of 2, and for 1,500 training steps. The optimizer and scheduler mirrored those used in the SFT phase. 
This standardized set-up ensured consistency and comparability across our experimental evaluations. All the implementation was done in python utilizing unsloth, huggingface, trl and vllm packages.

\subsection{ Loss Functions and Optimization}
The Supervised Fine-Tuning (SFT) process aims to minimize the standard auto-regressive language modeling loss (cross-entropy) over the reasoning trajectories. Let $\theta_{\text{base}}$ represent the frozen parameters of the base model and $\theta_{\text{LoRA}}$ represent the trainable LoRA adapter parameters. The SFT loss function is given by:

\begin{dmath}
\label{eq:sft_loss_app}
\mathcal{L}_{\text{SFT}}(\theta_{\text{LoRA}}) = - \sum_{j=1}^{|D_{\text{train}}|} \sum_{i=1}^{L_j} \log P(t_{j,i} | t_{j,1}, \dots, t_{j,i-1}; \theta_{\text{base}}, \theta_{\text{LoRA}})
\end{dmath}
where $D_{\text{train}}$ is the training set and $L_j$ is the length of the trajectory $T_j$.

For the GRPO phase, a new LoRA adapter was trained with weights $\phi$. The policy LLM is denoted as $\pi_{\phi}$. For each input prompt $x$ from the training dataset, the policy $\pi_{\phi}$ generates a group of $M$ distinct reasoning trajectories ${\tau^{(m)}}_{m=1}^M$. A binary reward $R(\tau) \in {0, 1}$ was assigned to each trajectory $\tau$ based on the exact match correctness of its final answer. The trajectory within the group of $M$ generations that achieved the highest reward (i.e., a correct answer, if any) was identified as $\tau^*$:

\begin{equation}
\label{eq:grpo_tau_star_app}
\tau^* = \operatorname*{argmax}_{\tau \in {\tau^{(m)}}_{m=1}^M} R(\tau)
\end{equation}

GRPO updates the policy by comparing the rewards of trajectories within each group. The optimization objective is to maximize the expected relative reward, which encourages the model to favor trajectories with higher relative rewards without relying on an explicit value function.

\subsection{Detailed Analysis of Reasoning Style Dynamics}
\label{subsec:appanalysis}
\begin{figure*}[t]
    \centering
    \includegraphics[width=0.75\linewidth]{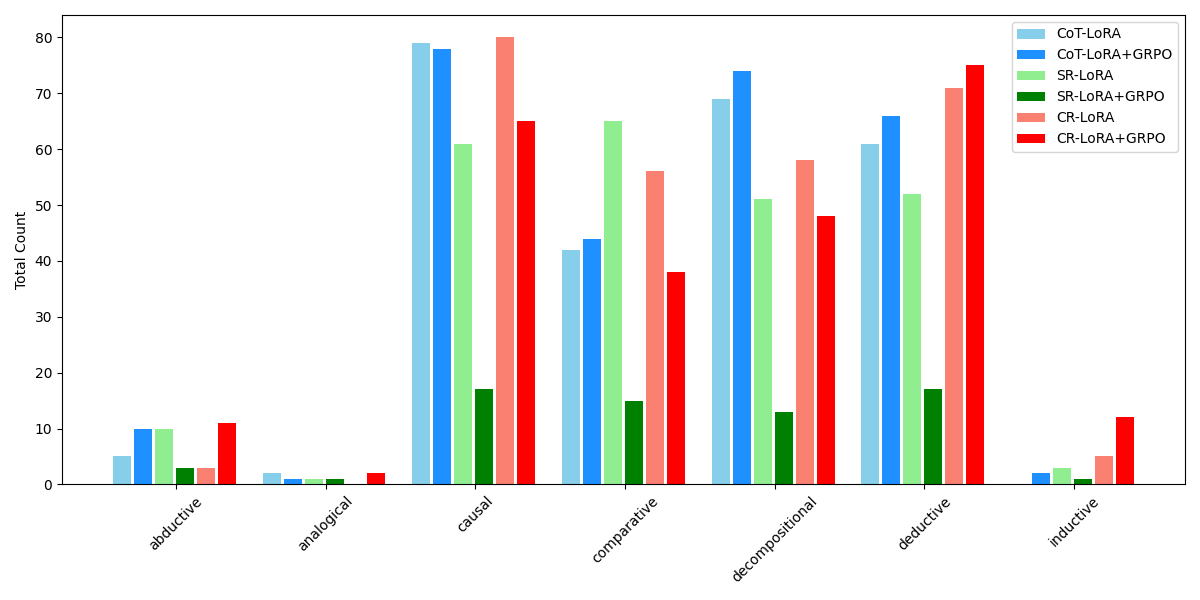}
    \caption{\scriptsize Analysis of reasoning strategies across the three trajectory types before and after applying GRPO tuning to the LoRA-tuned model on \textbf{ARC-C} dataset}
    \label{fig:arcanalysis}
\end{figure*}

\begin{figure*}
    \centering
    \includegraphics[width=0.75\linewidth]{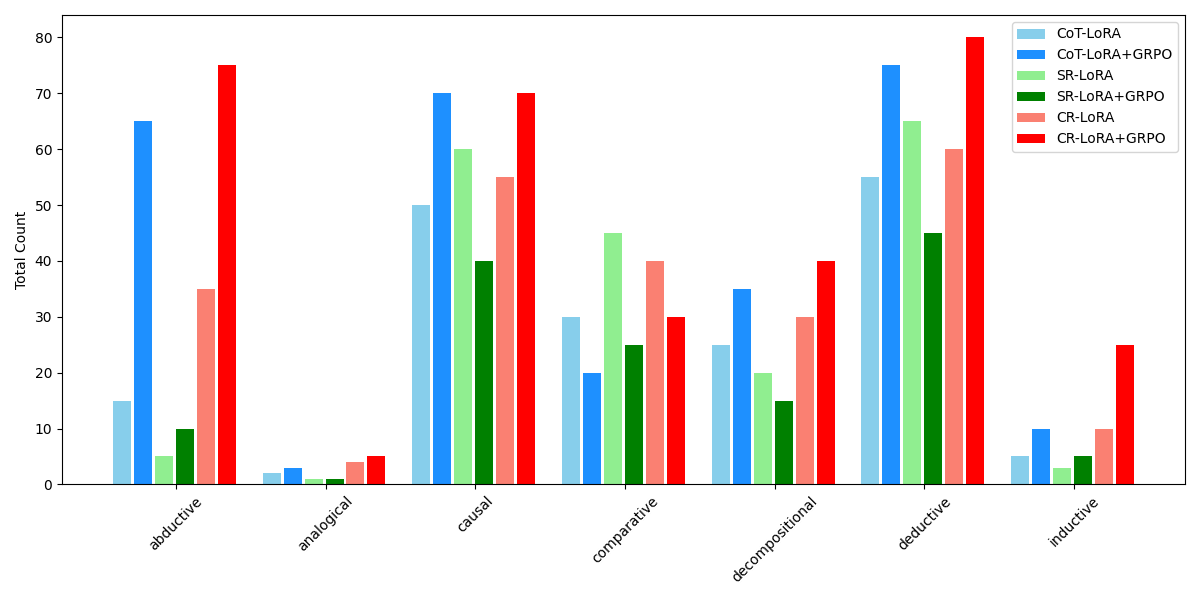}
    \caption{\scriptsize Analysis of reasoning strategies across the three trajectory types before and after applying GRPO tuning to the LoRA-tuned model on \textbf{MedMCQA} dataset}
    \label{fig:medanalysis}
\end{figure*}

The shift in reasoning style distribution due to GRPO, as depicted in the MedMCQA chart (Figure \ref{fig:medanalysis}) compared to the ARC-C chart (Figure \ref{fig:arcanalysis}), underscores how simply using outcome-based optimization adapts reasoning strategies to domain-specific demands, often showing a stylistic alignment with human cognitive approaches.

On MedMCQA—a domain demanding diagnostic inference—Composite Reasoning with GRPO (CR-LoRA+GRPO) markedly amplifies Abductive reasoning (inferring best explanations) and Deductive reasoning (applying medical rules), making them the dominant styles. By contrast, on ARC-C, GRPO’s CR primarily boosts Deductive and Causal reasoning, with only a modest uptick in Abductive and a stronger rise in Inductive reasoning. Chain-of-Thought post-GRPO on MedMCQA also increases Abductive and Deductive usage, but doesn’t reach the peaks achieved by CR. Likewise, on ARC-C, GRPO steers CR toward Causal, Deductive, Decompositional, and Inductive reasoning—reflecting the general science emphasis on cause-effect, logical breakdown, and generalization—while Abductive reasoning remains less prominent than in the medical setting.

The Standard Reasoning (SR) strategy, on MedMCQA, much like on ARC-C, shows a less adaptive pattern post-GRPO, with several of its initially high general reasoning styles (like Causal and Comparative) potentially decreasing or not being effectively channeled into medically critical styles like Abductive reasoning.

This domain-dependent adaptation is synergistic with human expert reasoning. Physicians often employ a hypothetico-deductive process, generating hypotheses (abduction) and testing them against evidence and knowledge (deduction) \cite{elstein1978medical}. The strong performance of the CR model and its post-GRPO reasoning profile in MedMCQA, with its emphasis on abductive and deductive styles, suggest that it learns to emulate these effective human diagnostic strategies more closely than other methods. Similarly, the broader scientific reasoning profile seen on ARC-C reflects the varied approaches humans use for general science problem-solving. The CR framework's flexibility, therefore, seems to allow GRPO to better identify and amplify the most effective, domain-appropriate human-like reasoning strategies.

%%%%%%%%%%%%%%%%%%%%%%%%%%%%%%%%%%%%%%%%%%%%%%%%%%%%%%%%%%%%

\end{document}